\documentclass[runningheads]{llncs}

\usepackage[T1]{fontenc}

\usepackage{graphicx}
\usepackage{listings}
\usepackage{cite}
\usepackage[colorlinks=true, allcolors=blue]{hyperref}
\usepackage{multirow}
\usepackage{amsmath}
\usepackage{amssymb}
\usepackage{mathtools}

\newcommand\Tstrut{\rule{0pt}{2.5ex}}         
\newcommand\Bstrut{\rule[-0.9ex]{0pt}{0pt}}   

\begin{document}
\title{Cross-domain Iterative Network for Simultaneous Denoising, Limited-angle Reconstruction, and Attenuation Correction of Low-dose Cardiac SPECT}
\titlerunning{Submission 709}


\author{Xiongchao Chen \and
Bo Zhou \and
Huidong Xie \and
Xueqi Guo \and
Qiong Liu \and
Albert J. Sinusas \and
Chi Liu}
\authorrunning{X. Chen et al.}

\institute{Yale University, New Haven, CT 06511, USA \\
\email{\{xiongchao.chen, chi.liu\}@yale.edu}}

\maketitle  

\begin{abstract}
Single-Photon Emission Computed Tomography (SPECT) is widely applied for the diagnosis of ischemic heart diseases. Low-dose (LD) SPECT aims to minimize radiation exposure but leads to increased image noise. Limited-angle (LA) SPECT enables faster scanning and reduced hardware costs but results in lower reconstruction accuracy. Additionally, computed tomography (CT)-derived attenuation maps ($\mu$-maps) are commonly used for SPECT attenuation correction (AC), but it will cause extra radiation exposure and SPECT-CT misalignments. In addition, the majority of SPECT scanners in the market are not hybrid SPECT/CT scanners. Although various deep learning methods have been introduced to separately address these limitations, the solution for simultaneously addressing these challenges still remains highly under-explored and challenging. To this end, we propose a Cross-domain Iterative Network (CDI-Net) for simultaneous denoising, LA reconstruction, and CT-free AC in cardiac SPECT. In CDI-Net, paired projection- and image-domain networks are end-to-end connected to fuse the emission and anatomical information across domains and iterations. Adaptive Weight Recalibrators (AWR) adjust the multi-channel input features to enhance prediction accuracy. Our experiments using clinical data showed that CDI-Net produced more accurate $\mu$-maps, projections, and reconstructions compared to existing approaches that addressed each task separately. Ablation studies demonstrated the significance of cross-domain and cross-iteration connections, as well as AWR, in improving the reconstruction performance. The source code is released at \href{https://**.com}{https://**.com}.

\end{abstract}

\section{Introduction}
Cardiac Single-Photon Emission Computed Tomography (SPECT) is the most widely performed non-invasive exam for clinical diagnosis of ischemic heart diseases \cite{danad2017comparison, gimelli2009stress}. Reducing the tracer dose can lower patient radiation exposure, but it will result in increased image noise \cite{wells2020dose, einstein2012effects}. Acquiring projections in fewer angles using fewer detectors allows for faster scanning and lower hardware costs, but it also leads to decreased reconstruction accuracy \cite{aghakhan2022deep, niu2014sparse}. Additionally, in clinical practice, computed tomography (CT)-derived attenuation maps ($\mu$-maps) are commonly used for SPECT attenuation correction (AC) \cite{blankespoor1996attenuation, goetze2007attenuation}. However, most SPECT scanners are stand-alone without the assistance of CT \cite{rahman2020fisher}. The CT scan also causes additional radiation exposure and SPECT-CT misalignments \cite{saleki2019influence, goetze2007attenuation}.


Deep learning-based methods have been extensively explored to address the aforementioned limitations individually. To reduce image noise in low-dose (LD) SPECT, convolutional neural networks (CNNs) were employed to process the LD projection, producing the full-dose (FD) projection for SPECT reconstruction \cite{sun2022deep, aghakhan2022deep}. Similarly, to perform limited-angle (LA) reconstruction, the LA projection was input to CNNs to predict the full-angle (FA) projection \cite{amirrashedi2021deep, whiteley2019cnn, shiri2020standard}. In addition, a dual-domain approach, known as Dual-domain Sinogram Synthesis (DuDoSS), utilized the image-domain output as a prior estimation for the projection domain to predict the FA projection \cite{chen2022dudoss}. For the CT-free AC, CNNs were used to generate pseudo attenuation maps ($\mu$-maps) from SPECT emission images \cite{shi2020deep, chen2022direct}. 


Although various methods have been developed to address these limitations individually, it is of great interest to address all these limitations simultaneously to enable CT-free, low-dose, low-cost, and accelerated SPECT, which could potentially lead to better performance on those separated but correlated tasks. Thus, we propose a Cross-Domain Iterative Network (CDI-Net) for simultaneous denoising, LA reconstruction, and CT-free AC in cardiac SPECT. In CDI-Net, projection and image-domain networks are end-to-end connected to fuse the predicted emission and anatomical features across domains and iterations. Adaptive Weight Recalibrators (AWR) calibrate the fused features to improve the prediction accuracy. We tested CDI-Net using clinical data and compared it to existing methods. Ablation studies were conducted to verify the impact of cross-domain, cross-iteration fusions and AWR on enhancing network performance.

\section{Materials and Methods}
\subsection{Problem Formulation}
The aim of this study is to generate the predicted FD$\&$FA projection ($\hat{P}_{F}$) and $\mu$-map ($\hat{\mu}$) with the LD$\&$LA projection (${P}_{L}$) as the inputs, formulated as:
\begin{equation}
   [ \hat{P}_{F}, \hat{\mu} ] = \mathcal{H} \left( {P}_{L}  \right), 
\end{equation}
where $\mathcal{H}\left( \cdot \right)$ is the pre-processing and neural network operator. The output labels are the ground-truth FD$\&$FA projection (${P}_{F}$) and CT-derived $\mu$-map (${\mu}$). Then, $\hat{P}_{F}$ and $\hat{\mu}$ are utilized in an offline maximum-likelihood expectation maximization (ML-EM, 30 iterations) module to reconstruct the target FD$\&$FA SPECT image with AC. Thus, predicting $\hat{P}_{F}$ performs the denoising and LA reconstruction, while predicting $\hat{\mu}$ enables the CT-free AC.

\begin{figure}[htb!]
\centering
\includegraphics[width=1.00\textwidth]{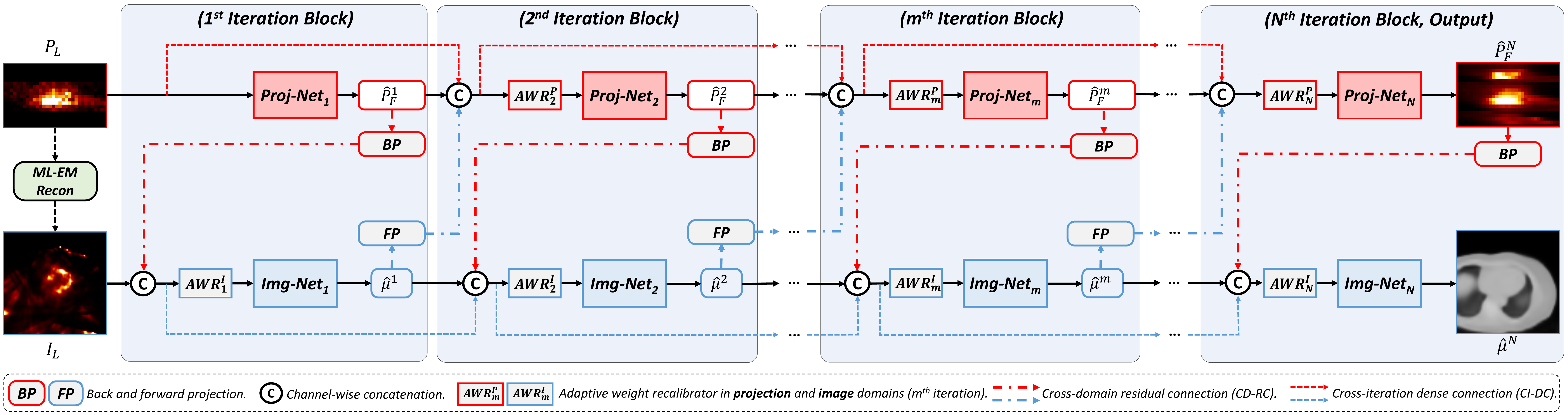}
\caption{Cross-Domain Iterative Networks. Projection- ($\textit{Proj-Net}$) and image-domain networks ($\textit{Img-Net}$) are end-to-end connected by cross-domain residual connections (CD-RC). Cross-iteration dense connections (CI-DC) enhance feature extraction across iterations. Adaptive Weight Recalibrators (AWR) adjust the multi-channel input.}
\label{fig:overview}
\end{figure}

\subsection{Dataset and Pre-processing}
This work includes 474 anonymized clinical hybrid SPECT/CT myocardial perfusion imaging (MPI) studies. Each study was conducted following the injection of $^{99\mathrm{m}}$Tc-tetrofosmin on a GE NM/CT 570c system. The GE 530c/570c system has 19 pinhole detectors placed in three columns on a cylindrical surface \cite{chan2016impact}. The clinical characteristics of the dataset are listed in supplementary Table S1. 

We extracted the 9 angles in the central column to simulate the configurations of the latest cost-effective MyoSPECT ES system \cite{gehealthcaremyospectessystems} as shown in supplementary Figure S1, generating the LA projection. The 10$\%$ LD projection was produced by randomly decimating the list-mode data using a 10$\%$ downsampling rate. ${P}_{L}$ was generated by combining the pipelines used for generating the LA and LD projections, and ${P}_{F}$ was the original FD$\&$FA projection. The ground-truth CT-derived $\mu$-maps ($\mu$) were well registered with the reconstructed SPECT images.

\subsection{Cross-Domain Iterative Network}
The overview framework of CDI-Net is shown in Fig.~\ref{fig:overview}. $P_L$ is first fed into an ML-EM reconstruction (30 iterations) module, producing the LD$\&$LA reconstructed SPECT image $I_L$, which is then employed for the $\mu$-map generation. \\

\noindent \textbf{Cross-Domain Residual Connection}. The projection- (\textit{Proj-Net}) and image-domain networks (\textit{Img-Net}) are both U-Net modules connected through cross-domain residual connections (\textbf{CD-RC}) facilitated by forward projection (FP) and back projection (BP) operators. In the 1$^{\mathrm{st}}$ iteration, $P_L$ is input to $\textit{Proj-Net}_1$ to generate $\hat{P}_F^1$ as:
\begin{equation}
    \hat{P}_F^1 = \mathcal{P}_1 \left( P_L \right),
\end{equation}
where $\mathcal{P}_1(\cdot)$ is $\textit{Proj-Net}_1$ operator. $\hat{P}_F^1$ is then processed by BP and introduced to $\textit{Img-Net}_1$ through CD-RC, providing emission information for the $\mu$-map generation. $I_L$ and the BP of $\hat{P}_F^1$ is first fed into the $\textit{AWR}^I_1$ (described in subsection 2.4) for multi-channel recalibration, and then input to $\textit{Img-Net}_1$ to generate $\hat{\mu}^1$:
\begin{equation}
    \hat{\mu}^1 = \mathcal{I}_1(\mathcal{A}_1^I (\left\{ {I}_{L}, \mathcal{T}_{b}(\hat{P}_F^1) \right\}) ),
\end{equation}
where $\mathcal{I}_1(\cdot)$ is the $\textit{Img-Net}_1$ operator. $\mathcal{A}_1^I(\cdot)$ refers to the $\textit{AWR}^I_1$ (superscript I means image-domain). $\left\{ \cdot \right\}$ is concatenation and $\mathcal{T}_{b}(\cdot)$ refers to BP. Next, the FP of $\hat{\mu}^1$ is added to $\textit{Proj-Net}_2$ of the next iteration by CD-RC, providing anatomical information for the projection prediction. This is the initial attempt of employing anatomical features for the estimation of FD$\&$FA projection in cardiac SPECT.

\begin{figure}[htb!]
\centering
\includegraphics[width=0.78\textwidth]{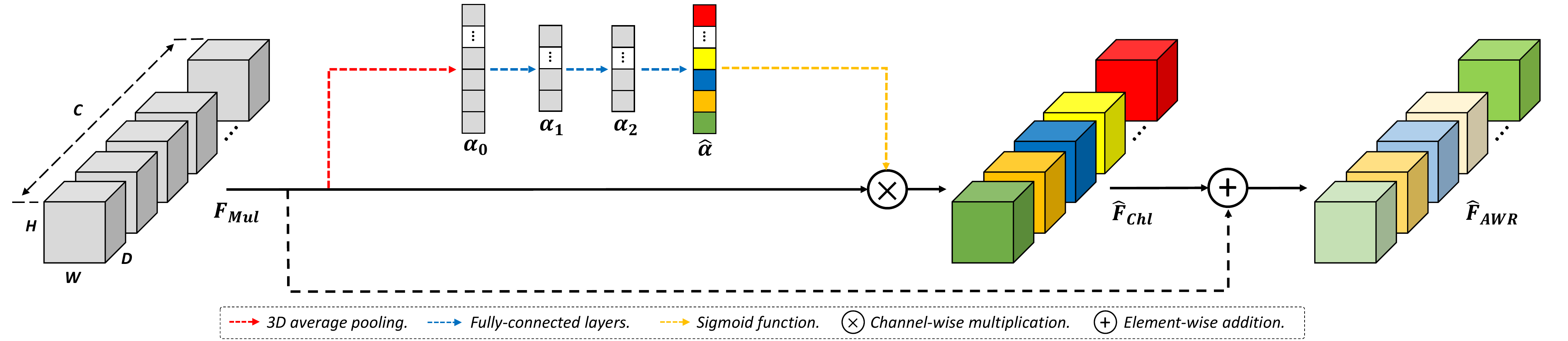}
\caption{Adaptive weight recalibrator. The channel weights of the input $F_{Mul}$ is first recalibrated. A global residual connection is then added to retain the original features.}
\label{fig:awr}
\end{figure}

\noindent \textbf{Cross-Iteration Dense Connection}. In the $m^{th} (m\geq2)$ iteration, the predicted projections from previous iterations, $\hat{P}_F^j (j<m)$, are incorporated into $\textit{Proj-Net}_{m}$ through cross-iteration dense connections (\textbf{CI-DC}). The FP of $\hat{\mu}^j$ from $\textit{Img-Net}_j (j<m)$ are also added to $\textit{Proj-Net}_{m}$ through CD-RC as additional input anatomical information. The multi-channel input of $\textit{Proj-Net}_{m}$ is:
\begin{equation}
\label{eq:input_proj}
    U_{P}^{m} = \left\{P_L,  \hat{P}_F^1, \hat{P}_F^2,  \cdots, \hat{P}_F^{(m-1)}, \mathcal{T}_{f}(\hat{\mu}^{1}), \mathcal{T}_{f}(\hat{\mu}^{2}), \cdots, \mathcal{T}_{f}(\hat{\mu}^{(m-1)}) \right\},
\end{equation}
where $\mathcal{T}_{f}(\cdot)$ refers to the FP. Then, $U_{P}^{m}$ is fed into $\textit{AWR}^P_m$ for recalibration and input to $\textit{Proj-Net}_{m}$ to generate $\hat{P}_F^m$, formulated as:
\begin{equation}
    \hat{P}_F^m = \mathcal{P}_m(\mathcal{A}_m^P (U_{P}^{m})),
\end{equation}
where $\mathcal{P}_m(\cdot)$ refers to $\textit{Proj-Net}_m$ and $\mathcal{A}_m^P(\cdot)$ is $\textit{AWR}^P_m$. Similarly, the predicted $\mu$-maps from previous iterations, $\hat{\mu}^{j} (j<m)$, are integrated into $\textit{Img-Net}_{m}$ by CI-DC. The BP of $\hat{P}_F^j (j \leq m)$ are also added to $\textit{Img-Net}_{m}$ by CD-RC as additional input emission information. The multi-channel input of $\textit{Img-Net}_{m}$ is:
\begin{equation}
\label{eq:input_img}
    U_{I}^{m} = \left\{ {I}_{L}, \hat{\mu}^{1}, \hat{\mu}^{2}, \cdots, \hat{\mu}^{(m-1)}, \mathcal{T}_{b}(\hat{P}_F^1), \mathcal{T}_{b}(\hat{P}_F^2), \cdots, \mathcal{T}_{b}(\hat{P}_F^{(m-1)}), \mathcal{T}_{b}(\hat{P}_F^m)      \right\}.
\end{equation}
Then, $U_{I}^{m}$ is recalibrated by $\textit{AWR}^I_m$ and input to $\textit{Img-Net}_{m}$ to produce $\hat{\mu}^m$ as:
\begin{equation}
    \hat{\mu}^m = \mathcal{I}_m(\mathcal{A}_m^I ( U_{I}^{m})),
\end{equation}
where $\mathcal{I}_m(\cdot)$ is $\textit{Img-Net}_m$ and $\mathcal{A}_m^I(\cdot)$ is the $\textit{AWR}^I_m$ operator.   \\

\noindent \textbf{Loss function}. The network outputs are $\hat{P}_F^N$ and $\hat{\mu}^N$, where $N$ is the number of iterations (default: 5). The overall loss function $\mathcal{L}$ is formulated as:
\begin{equation}
    \mathcal{L} = \sum_{i=1}^{N}( w_P \left\| \hat{P}^{i}_F - P_F  \right\|_1   +  w_{\mu} \left\| \hat{\mu}^i - \mu  \right\|_1),
\end{equation}
where $w_P$ and $w_{\mu}$ are the weights of the projection- and image-domain losses. In our experiment, we empirically set $w_P=0.5$ and $w_{\mu}=0.5$ for balanced training.

\subsection{Adaptive Weight Recalibrator}
The diagram of AWR is shown in Fig.~\ref{fig:awr}. As presented in Eq. \ref{eq:input_proj} and \ref{eq:input_img}, the multi-channel input consists of emission and anatomical features, formulated as:
\begin{equation}
    F_{Mul} = [f_1, f_2, \dots, f_C],
\end{equation}
where $f_i \in \mathbb{R}^{H \times W \times D}$ indicates the emission or anatomical feature in each individual channel. $F_{Mul}$ is flattened using 3D average pooling, producing $\alpha_0$ that embeds the channel weights. A recalibration vector $\hat{\alpha}$ is generated using fully-connected layers and a Sigmoid function. $\hat{\alpha}$ is applied to $F_{Mul}$, described as:
\begin{equation}
    \hat{F}_{Chl} =  [f_1 \hat{\alpha}_1, f_2 \hat{\alpha}_2, \dots, f_C \hat{\alpha}_C],
\end{equation}
where $\hat{\alpha}_i \in [0, 1]$ indicates the channel recalibration factor. Then, a global residual connection is applied to retain the original information, producing the output of AWR as $\hat{F}_{AWR} = \hat{F}_{Chl} + F_{Mul}$. Thus, AWR adaptively adjusts the weight of each input channel to better integrate the emission and anatomical information for higher prediction accuracy.

\begin{figure}[htb!]
\centering
\includegraphics[width=0.88\textwidth]{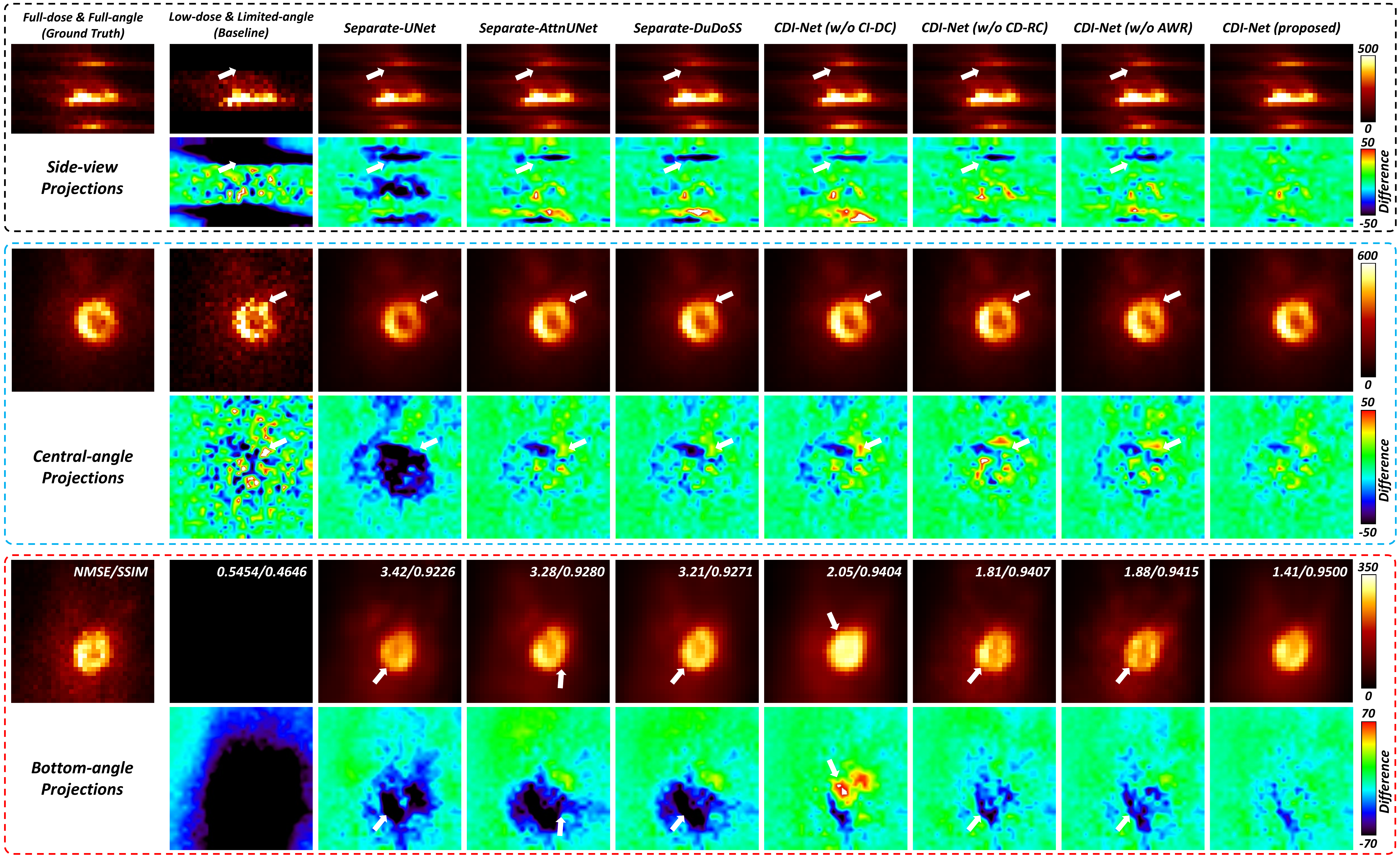}
\caption{Predicted FD$\&$FA projections in side, central-angle, and bottom-angle views with NMSE/SSIM annotated. White arrows point out the prediction inconsistency.}
\label{fig:proj}
\end{figure}

\subsection{Implementation Details}
In this study, we tested CDI-Net against many existing methods in terms of the predicted FD$\&$FA projections, $\mu$-maps, and AC SPECT images. U-Net (labeled as Separate-UNet) \cite{ronneberger2015u}, Attention U-Net (labeled as Separate-AttnUNet) \cite{oktay2018attention}, and DuDoSS (labeled as Separate-DuDoSS) \cite{chen2022dudoss} were applied to generate $\hat{P}_F$ with $P_L$ as input. U-Net and Attention U-Net were also employed to predict $\hat{\mu}$ with $I_L$ as input. We also tested ablation study groups w/o CI-DC, CD-RC, or AWR. Then, $\hat{P}_F$ and $\hat{\mu}$ were utilized to reconstruct the AC images. 

All networks were developed using PyTorch \cite{paszke2019pytorch} with Adam optimizers \cite{kingma2014adam}. The image- and projection-domain modules were trained with initial learning rates (LR) of $10^{-3}$ and $10^{-4}$ with a decay rate of 0.99/epoch \cite{you2019does}. The networks that predict $\mu$-maps or projections separately were trained for 200 epochs, while CDI-Net was trained for 50 epochs. The performance of CDI-Net with different iterations (1 to 6, default 5) is presented in section 3 (Fig.~\ref{fig:plot}), and the impact of multiple LD levels (1 to 80$\%$, default 10$\%$) is shown in section 3 (Fig.~\ref{fig:plot}).

\begin{table} [htb!]
\caption{Evaluation of the predicted projections using normalized mean square error (NMSE), normalized mean absolute error (NMAE), structural similarity (SSIM), and peak signal-to-noise ratio (PSNR). The best results are marked in \textcolor{red}{red}.}
\label{tab:proj} 
\tiny
\centering
\begin{tabular}{ l | c | c | c | c || c}

\hline
\textbf{Methods}             & \textbf{NMSE($\boldsymbol{\%}$)}     & \textbf{NMAE($\boldsymbol{\%}$)}    & \textbf{SSIM}                   & \textbf{PSNR}                & \textbf{P-values$^{\dag}$}   \Tstrut\Bstrut\\  
\hline  
Baseline LD$\&$LA       & $54.56 \pm 2.46$                     & $62.44 \pm 2.53$                    & $0.4912 \pm 0.0260$             & $19.23 \pm 1.68$             & < 0.001                      \Tstrut\Bstrut\\
\hline
Separate-UNet \cite{ronneberger2015u}    & $4.21 \pm 1.48$                      & $16.69 \pm 2.24$                    & $0.9276 \pm 0.0195$             & $30.58 \pm 1.79$             & < 0.001                      \Tstrut\Bstrut\\  
Separate-AttnUNet \cite{oktay2018attention}      & $3.45 \pm 1.13$        & $15.45 \pm 2.56$                    & $0.9368 \pm 0.0205$             & $31.43 \pm 1.65$             & < 0.001                      \Tstrut\Bstrut\\  
Separate-DuDoSS \cite{chen2022dudoss}    & $3.19 \pm 1.11$                      & $14.57 \pm 2.29$                    & $0.9416 \pm 0.0187$             & $31.79 \pm 1.65$             & < 0.001                      \Tstrut\Bstrut\\  
CDI-Net (w/o CI-DC)    & $2.56 \pm 0.85$                      & $13.22 \pm 1.81$                    & $0.9505 \pm 0.0144$             & $32.73 \pm 1.65$             & < 0.001                      \Tstrut\Bstrut\\  
CDI-Net (w/o CD-RC)    & $2.39 \pm 0.78$                      & $13.39 \pm 1.94$                    & $0.9486 \pm 0.0160$             & $33.02 \pm 1.65$             & < 0.001                      \Tstrut\Bstrut\\  
CDI-Net (w/o AWR)     & $2.42 \pm 0.83$                      & $13.40 \pm 2.00$                    & $0.9478 \pm 0.0173$             & $32.98 \pm 1.65$             & < 0.001                      \Tstrut\Bstrut\\  
CDI-Net (proposed)      & \textcolor{red}{$2.15 \pm 0.69$}             & \textcolor{red}{$12.64 \pm 1.77$}           & \textcolor{red}{$0.9542 \pm 0.0142$}    & \textcolor{red}{$33.47 \pm 1.68$}    & \textendash                  \Tstrut\Bstrut\\  
\hline
\multicolumn{6}{l}{$^{\dag}$P-values of the paired t-tests of NMSE between the current method and CDI-Net (proposed).} \\
\end{tabular}
\end{table}

\section{Results}
Fig.~\ref{fig:proj} shows the predicted FD$\&$FA projections in multiple views. We can observe that CDI-Net outputs more accurate projections than other groups. Conversely, Separate-UNet, Separate-AttnUNet, and Separate-DuDoSS display underestimations in cardiac regions. This indicates the advantages of fusing emission and anatomical features for simultaneous prediction as in the CDI-Net. Moreover, CDI-Net shows superior performance to the ablation study groups w/o CI-DC, CD-RC, or AWR, confirming the significance of CI-DC, CD-RC, and AWR in enhancing network performance. Table~\ref{tab:proj} lists the quantitative comparison of the predicted projections. CDI-Net produces more accurate quantitative results than groups conducting separate predictions and ablation study groups ($p<0.001$).

\begin{figure}[htb!]
\centering
\includegraphics[width=0.75\textwidth]{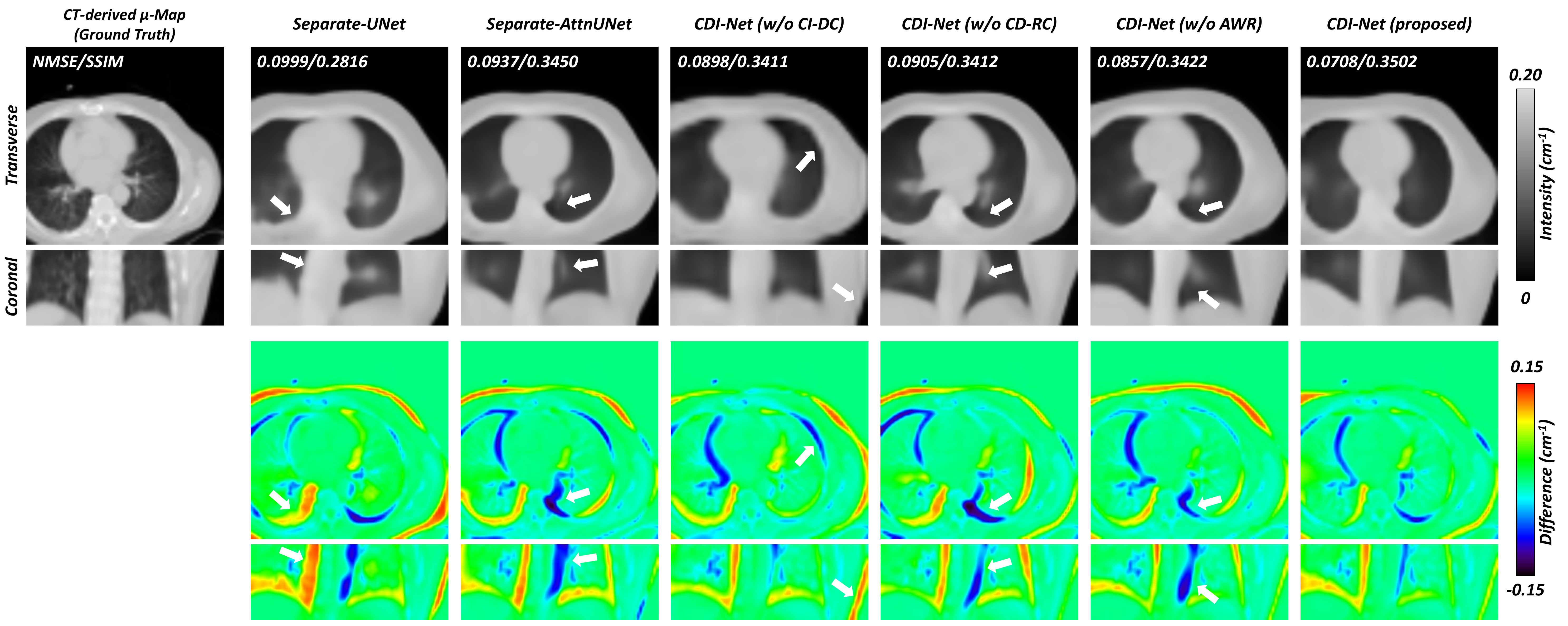}
\caption{Predicted $\mu$-maps. White arrows denote the prediction inconsistency.}
\label{fig:amap}
\end{figure}

\begin{table} [htb!]
\caption{Evaluation of the predicted $\mu$-maps. The best results are marked in \textcolor{red}{red}.}
\label{tab:amap} 
\tiny
\centering
\begin{tabular}{ l | c | c | c | c || c}

\hline
\textbf{Methods}               & \textbf{NMSE($\boldsymbol{\%}$)}     & \textbf{NMAE($\boldsymbol{\%}$)}    & \textbf{SSIM}                   & \textbf{PSNR}                & \textbf{P-values$^{\dag}$}    \Tstrut\Bstrut\\  
\hline
Separate-UNet  \cite{ronneberger2015u}   & $12.83 \pm 4.85$                      & $22.93 \pm 5.96$                    & $0.2782 \pm 0.0617$             & $17.22 \pm 1.81$             & < 0.001                      \Tstrut\Bstrut\\  
Separate-AttnUNet \cite{oktay2018attention}       & $12.45 \pm 4.49$                      & $22.20 \pm 5.49$                    & $0.2829 \pm 0.0582$             & $17.34 \pm 1.82$           & < 0.001         \Tstrut\Bstrut\\  
CDI-Net (w/o CI-DC)    & $11.88 \pm 4.18$                      & $22.69 \pm 5.37$                    & $0.2993 \pm 0.0624$             & $17.54 \pm 1.81$             & < 0.001                      \Tstrut\Bstrut\\  
CDI-Net (w/o CD-RC)    & $11.90 \pm 4.69$                      & $21.95 \pm 5.60$                    & $0.3041 \pm 0.0660$             & $17.56 \pm 1.80$             & < 0.001                      \Tstrut\Bstrut\\  
CDI-Net (w/o AWR)     & $11.84 \pm 4.69$                      & $21.96 \pm 5.48$                    & $0.3047 \pm 0.0627$             & $17.60 \pm 1.89$             & < 0.001                      \Tstrut\Bstrut\\  
CDI-Net (proposed)    & \textcolor{red}{$11.42 \pm 4.31$}             & \textcolor{red}{$21.54 \pm 5.30$}           & \textcolor{red}{$0.3066 \pm 0.0607$}    & \textcolor{red}{$17.83 \pm 1.85$}    & \textendash                  \Tstrut\Bstrut\\  
\hline
\multicolumn{6}{l}{$^{\dag}$P-values of the paired t-tests of NMSE between the current method and CDI-Net (proposed).} \\
\end{tabular}
\end{table}

Fig.~\ref{fig:amap} presents the predicted $\mu$-maps. It can be observed that CDI-Net outputs more accurate $\mu$-maps than other testing groups. The $\mu$-maps predicted by Separate-UNet and Separate-AttnUNet display obvious inconsistency with the ground truth, particularly in the inner boundaries. This indicates that CDI-Net improves the accuracy of generating $\mu$-maps by incorporating emission and anatomical information. Moreover, the $\mu$-map predicted by CDI-Net is more accurate than ablation study groups w/o CI-DC, CD-RC, or AWR. Table~\ref{tab:amap} lists the quantitative evaluation of the predicted $\mu$-maps. The $\mu$-maps predicted by CDI-Net exhibit lower quantitative errors than other methods ($p<0.001$).

\begin{figure}[htb!]
\centering
\includegraphics[width=0.85\textwidth]{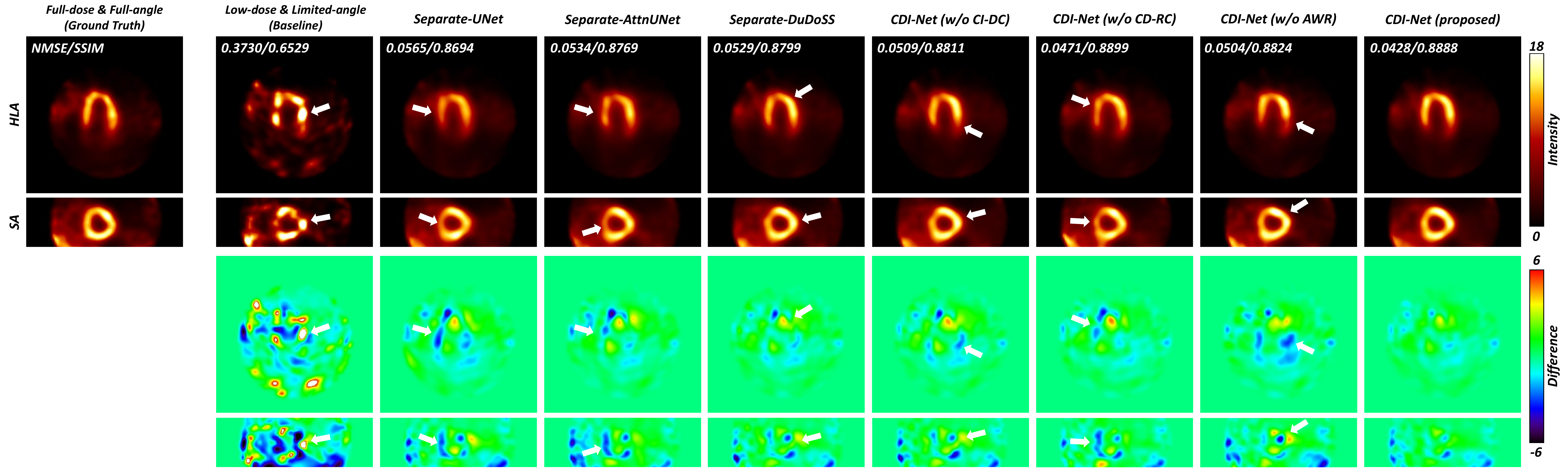}
\caption{Reconstructed SPECT images with attenuation correction (AC) using the predicted FD$\&$FA projections and $\mu$-maps. White arrows denote the inconsistency.}
\label{fig:recon}
\end{figure}

The predicted projections and $\mu$-maps are then utilized in SPECT reconstruction. As shown in Fig.~\ref{fig:recon}, CDI-Net produces the most accurate AC images. The groups conducting separate predictions or ablation study groups show over- or under-estimations of the myocardial perfusion intensities compared to the ground truth. The quantitative evaluation listed in Table~\ref{tab:recon} shows that CDI-Net leads to higher reconstruction accuracy than other testing groups ($p<0.001$). Segment-wise evaluation of the AC images is shown in supplementary Fig. S2.

\begin{table} [htb!]
\caption{Evaluation of the reconstructed AC SPECT. The best results are in \textcolor{red}{red}.}
\label{tab:recon} 
\tiny
\centering
\begin{tabular}{ l | c | c | c | c || c}

\hline
\textbf{Methods}             & \textbf{NMSE($\boldsymbol{\%}$)}     & \textbf{NMAE($\boldsymbol{\%}$)}    & \textbf{SSIM}                   & \textbf{PSNR}                & \textbf{P-values$^{\dag}$}   \Tstrut\Bstrut\\  
\hline  
Baseline LD$\&$LA     & $35.80 \pm 10.83$                    & $54.36 \pm 6.13$                    & $0.6646 \pm 0.0344$             & $24.00 \pm 1.80$             & < 0.001                      \Tstrut\Bstrut\\
\hline
Separate-UNet  \cite{ronneberger2015u}      & $6.63 \pm 2.26$                      & $23.78 \pm 3.53$                    & $0.8576 \pm 0.0248$             & $31.33 \pm 1.72$             & < 0.001              \Tstrut\Bstrut\\  
Separate-AttnUNet \cite{oktay2018attention}    & $5.85 \pm 1.76$                      & $22.46 \pm 2.96$                    & $0.8655 \pm 0.0239$             & $31.56 \pm 1.67$             & < 0.001        \Tstrut\Bstrut\\  
Separate-DuDoSS \cite{chen2022dudoss}   & $5.68 \pm 1.81$                      & $22.02 \pm 3.11$                    & $0.8706 \pm 0.0242$             & $32.00 \pm 1.70$             & < 0.001                      \Tstrut\Bstrut\\  
CDI-Net (w/o CI-DC)    & $5.45 \pm 1.61$                      & $21.67 \pm 2.92$                    & $0.8742 \pm 0.0207$             & $32.15 \pm 1.69$             & < 0.001                      \Tstrut\Bstrut\\  
CDI-Net (w/o CD-RC)    & $5.55 \pm 1.81$                      & $21.66 \pm 3.13$                    & $0.8722 \pm 0.0231$             & $32.12 \pm 1.69$             & < 0.001                      \Tstrut\Bstrut\\  
CDI-Net (w/o AWR)     & $5.49 \pm 1.66$                      & $21.59 \pm 2.92$                    & $0.8729 \pm 0.0224$             & $32.13 \pm 1.70$             & < 0.001                      \Tstrut\Bstrut\\  
CDI-Net (proposed)              & \textcolor{red}{$4.82 \pm 1.44$}             & \textcolor{red}{$20.28 \pm 2.65$}           & \textcolor{red}{$0.8829 \pm 0.0194$}    & \textcolor{red}{$32.69 \pm 1.65$}    & \textendash                  \Tstrut\Bstrut\\  
\hline
\multicolumn{6}{l}{$^{\dag}$P-values of the paired t-tests of NMSE between the current method and CDI-Net (proposed).} \\
\end{tabular}
\end{table}

\begin{figure}[htb!]
\centering
\includegraphics[width=0.88\textwidth]{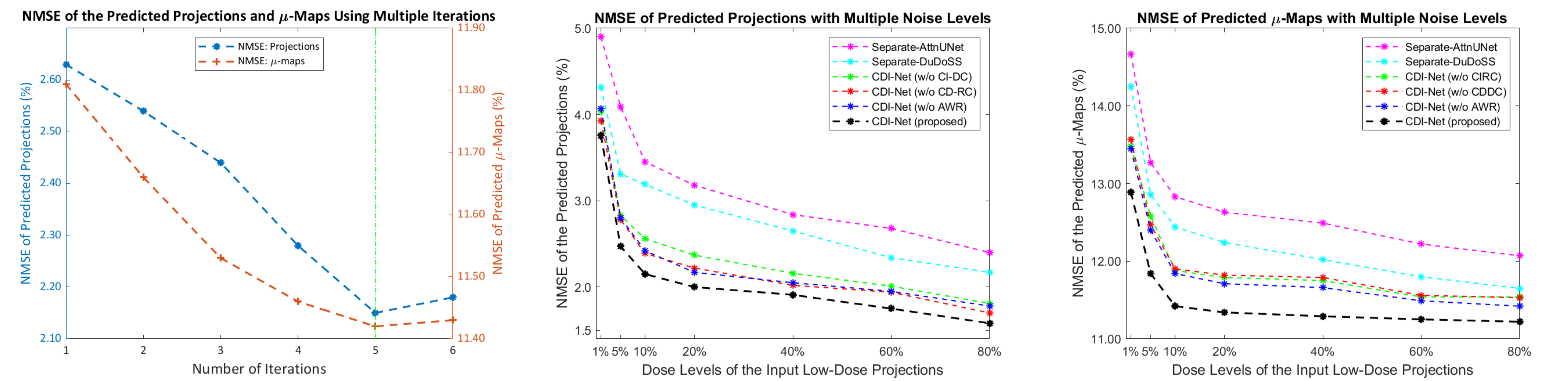}
\caption{Evaluation of CDI-Net with different iterations (left). Evaluation of multiple methods based on datasets with multiple low-dose levels (mid, right).}
\label{fig:plot}
\end{figure}

Moreover, we tested the performance of CDI-Net with different iterations as shown in Fig.~\ref{fig:plot} (left). The errors of the predicted projections and $\mu$-maps by CDI-Net decrease as the number of iterations increases, with convergence occurring at 5 iterations. Additionally, we generated more datasets with multiple LD levels to test these methods as shown in Fig.~\ref{fig:plot} (mid, right). It can be observed that CDI-Net demonstrates consistently higher prediction accuracy of projections and $\mu$-maps than other groups across multiple LD levels.

\section{Discussion and Conclusion}
In this paper, we propose CDI-Net that simultaneously achieves denoising, LA reconstruction, and CT-free AC for low-dose cardiac SPECT. The CD-RC and CI-DC components effectively fuse the predicted anatomical and emission features. The fused features are adaptively calibrated by AWR and then jointly employed for the prediction of projections and $\mu$-maps. Thus, CDI-Net effectively combines the cross-domain information that is then used for image estimations in both domains. This approach also marks the initial investigation in employing anatomical features to assist the projection estimation of cardiac SPECT. Experiments using clinical data with different LD levels demonstrated the superiority of CDI-Net over existing methods in predicting projections and $\mu$-maps, as well as in reconstructing AC SPECT images. 

For potential clinical impact, CDI-Net enables accurate AC SPECT reconstruction in LD, LA, and CT-less scenarios. This could potentially promote the clinical adoption of the latest cost-effective SPECT scanners with fewer detectors and lower dose levels and without CT. Thus, we can achieve accurate cardiac AC SPECT imaging with reduced hardware expenses and lower radiation exposure.

\bibliographystyle{splncs04}
\bibliography{reference}

\newpage
\setcounter{section}{0}
\setcounter{figure}{0}
\setcounter{table}{0}
\section{Supplementary Information}

\subsection{Clinical characteristics of the patients in the dataset}

\begin{table} [htb!]
\caption{(SI) The gender, height, weight, and BMI distributions of the enrolled patients.}
\label{tab:data} 
\scriptsize
\centering
\begin{tabular}{ l | c | c | c | c | c}
\hline
\multicolumn{2}{l|}{\textbf{Datasets}}                          & \textbf{Age (year)}  & \textbf{Height (m)}  & \textbf{Weight (kg)}  & \textbf{BMI}      \Tstrut\Bstrut\\
\hline
\multirow{2}{*}{Training (108 M$^{\dag}$, 92 F$^{\ddag}$)}  & Range                & $27 - 86$            & $1.32 - 2.03$        & $44.91 - 127.00$      & $18.10 - 48.05$   \Tstrut\Bstrut\\
                                         & Mean $\pm$ Std.      & $65.0 \pm 11.6$      & $1.68 \pm 0.11$      & $85.67 \pm 20.62$     & $30.29 \pm 6.57$  \Tstrut\Bstrut\\
\hline
\multirow{2}{*}{Validation (52 M, 22 F)} & Range                & $41 - 84$            & $1.47 - 1.85$        & $54.34 - 103.87$      & $19.53 - 38.11$   \Tstrut\Bstrut\\
                                         & Mean $\pm$ Std.      & $65.5 \pm 10.1$      & $1.70 \pm 0.09$      & $81.28 \pm 12.14$     & $28.33 \pm 4.43$  \Tstrut\Bstrut\\
\hline
\multirow{2}{*}{Testing (104 M, 96 F)}   & Range                & $39 - 87$            & $1.47 - 1.98$        & $45.00 - 140.00$      & $18.26 - 48.44$   \Tstrut\Bstrut\\
                                         & Mean $\pm$ Std.      & $64.2 \pm 10.7$      & $1.69 \pm 0.11$      & $86.41 \pm 18.91$     & $30.28 \pm 5.81$  \Tstrut\Bstrut\\
\hline  
\multicolumn{4}{l}{$^{\dag}$M stands for male.} \Tstrut\Bstrut\\
\multicolumn{4}{l}{$^{\ddag}$F stands for female.} \Tstrut\Bstrut\\
\end{tabular}
\end{table}

\subsection{Latest SPECT system with fewer detectors}
\begin{figure}[htb!]
\centering
\includegraphics[width=1.00\textwidth]{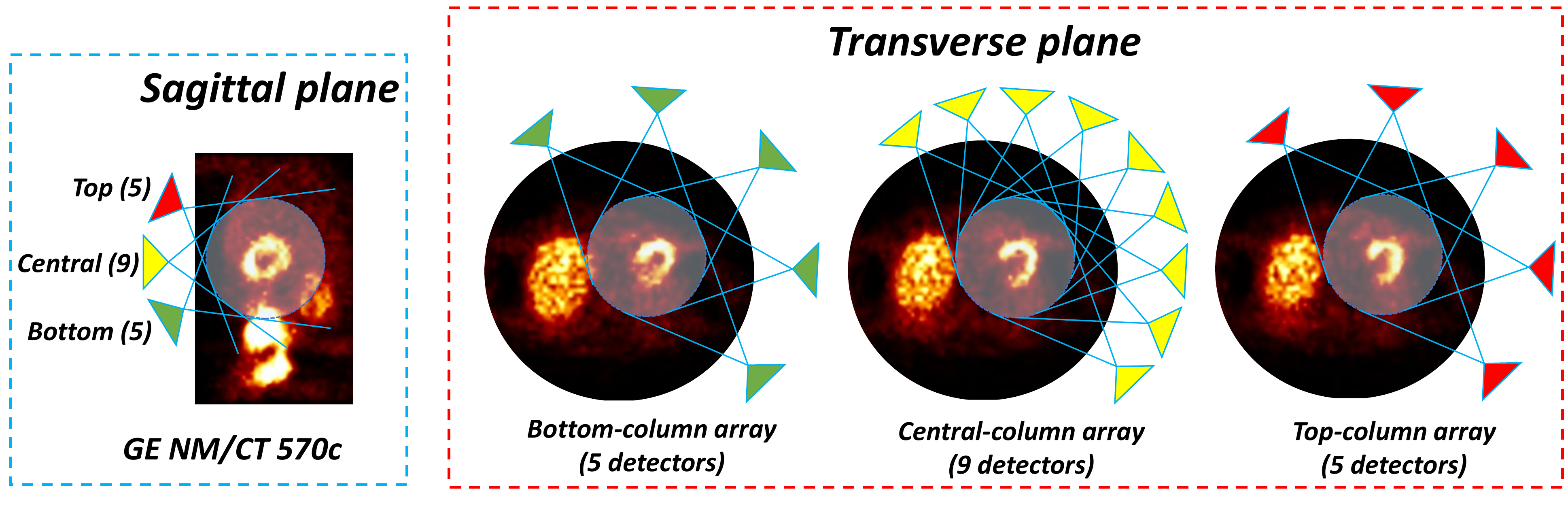}
\caption{(SI) Configurations of GE NM/CT 530c/570c and the latest SPECT system with fewer detectors. GE NM/CT 570c scanner comprises of 19 pinhole detectors arranged in three columns on a cylindrical surface (left blue box) with 5, 9, 5 detectors placed on bottom, central, and top columns, respectively (right red box). The most recent few-angle scanner only employ the 9 detectors at the central column for minimizing hardware expenses.}
\label{fig:configuration}
\end{figure}

\newpage
\subsection{Segment-wise evaluations of the AC SPECT images}
\begin{figure}[htb!]
\centering
\includegraphics[width=0.88\textwidth]{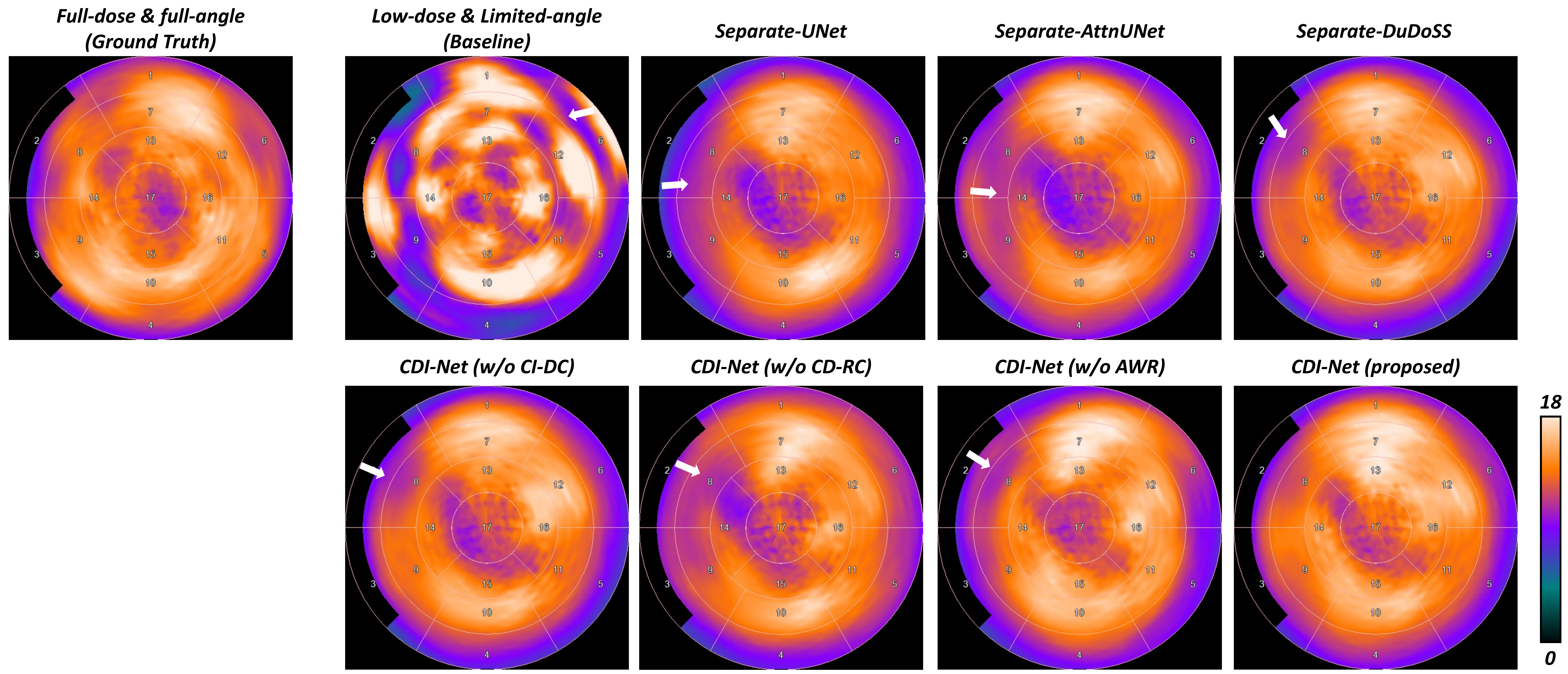}
\caption{(SI) The AC SPECT images were analyzed using standard 17-segment polar maps, with white arrows indicating prediction inconsistencies. CDI-Net produces the most accurate polar maps, while the groups conducting separate prediction and ablation study groups display over- or underestimation of the segment intensities.}
\label{fig:polarmap}
\end{figure}

\end{document}